\documentclass[runningheads]{llncs}
\usepackage[UTF8, scheme=plain]{ctex}
\usepackage[T1]{fontenc}
\usepackage{amsmath,amssymb}
\usepackage{graphicx}
\usepackage{subcaption}
\usepackage{booktabs}
\usepackage{hyperref}
\begin{document}
%
\title{Operator Fusion for LLM Inference on the Tensix Architecture}
%
%
\author{Qingbo Wu\inst{1} \and
Ke Li\inst{2} \and
Wenzhu Wang\inst{2} \and
Jie Yu\inst{3} \and
Ruian Zhang\inst{2} \and
Lili Liu\inst{2}}
\authorrunning{Q. Wu et al.}
%
\institute{KylinSoft Co., Ltd, Tianjin, China \and
openKylin Community, Tianjin, China\\
\email{li\_ke@hl-it.cn} \and
College of Computer Science and Technology, National University of Defense Technology, Changsha, China}
\maketitle              
\begin{abstract}
\narrower
This study addresses on-device inference bottlenecks of Transformer models on Tenstorrent's Tensix architecture and proposes an operator fusion strategy that enhances data locality. RMSNorm is fused with matrix multiplication in self-attention and in the FFN, enabling back-to-back execution of memory-bound and compute-bound operators in on-chip SRAM to significantly reduce DRAM reads/writes of intermediate results and scheduling overhead. 
To support multi-core parallelism, a NoC-based multicast mechanism is leveraged in which row/column master nodes efficiently distribute inputs and weights across the core mesh, alleviating DRAM bandwidth contention. Experiments on the Wormhole platform with Qwen2.5-0.5B, Qwen3-0.6B, and Qwen3-4B show up to 37.44\% latency reduction for attention and 15.89\% for MLP, with up to 7.91\% reduction per decoder layer, while Pearson Correlation Coefficient (PCC) remains above 98.75\%, confirming significant end-to-end efficiency gains under numerical consistency.
\par
\end{abstract}
\section{Introduction}

As large language models are widely applied in code generation, QA, and multimodal reasoning, performance and energy efficiency for edge inference have become increasingly critical \cite{yuan2024llm}. During inference, Transformer workloads exhibit a pronounced “read-compute-write” pattern: memory-bound operators (normalization, activation) alternate with compute-bound operators (matrix multiplications). This causes frequent movement of intermediate results between off-chip DRAM and on-chip SRAM, incurring high memory latency and scheduling overhead. On resource-constrained devices (e.g., Tenstorrent Tensix), improving data locality, reducing redundant data movement, and overlapping compute with memory accesses are key to efficient inference.

We propose a dataflow- and memory-hierarchy-aware operator fusion strategy tailored to decoder-only Transformers and constrained by Tensix architecture. Specifically, we fuse RMSNorm with the QKV projection matmuls in self-attention, and fuse RMSNorm with the first FFN matmul in the second residual block. These are executed back-to-back in on-chip SRAM to avoid DRAM writebacks and subsequent reads. For multi-core parallelism, we leverage a NoC-based row/column multicast mechanism in which master nodes efficiently distribute inputs and weights across the core mesh, alleviating DRAM bandwidth contention and improving memory efficiency.

Our experiments use Tenstorrent's Tensix architecture and the Wormhole N300 accelerator as the evaluation platform \cite{Wormhole18:online}. The Tensix processor employs a 2D mesh interconnect and emphasizes decoupling compute, control, and communication. Each core integrates multiple RISC\textendash V control cores, specialized FPU/SFPU compute units, and large local SRAM, enabling on-chip dataflow and operator-level parallelism \cite{tenstorr70:online}. Wormhole N300 integrates two Tensix chips and can flexibly switch between 64/128 core configurations, providing a reconfigurable hardware substrate and a realistic evaluation environment for the proposed fusion and multicast mechanisms.

The main contributions of this study are as follows:
\begin{itemize}
\item Propose a single-core fusion scheme for Transformers: fuse RMSNorm with QKV/FFN matmuls in a pass-through manner within a Tensix core to maximize locality and reduce scheduling overhead.
\item Adopt a 2D mesh mapping and leverage the NoC multicast mechanism for multi-core execution, exploiting row/column data sharing to reduce bandwidth pressure on the Tensix core mesh.
\item Conduct a systematic evaluation on the Tenstorrent Wormhole N300 with Qwen2.5-0.5B, Qwen3-0.6B, and Qwen3-4B. Results show substantial latency reductions for key modules and the decoder layer while maintaining numerical consistency (PCC ≥ 98.75\%).
\end{itemize}

\section{Related Work}

\subsection{Large Language Models}

Modern large language models (LLMs) predominantly adopt the decoder-only architecture. A single decoder layer typically contains two residual modules: the self-attention module and the feed-forward network (FFN). Both modules center around matrix multiplications: self-attention performs linear projections to produce Query, Key, and Value vectors as well as the final output projection; the FFN comprises multiple linear layers (e.g., SwiGLU in Llama includes Gate, Up, and Down projection matrices) to perform nonlinear transformations and dimensional mappings. At the input of these modules, RMSNorm is commonly used as a normalization layer to improve training stability.

\begin{figure}[htbp]
\centering
\includegraphics[width=0.4\textwidth]{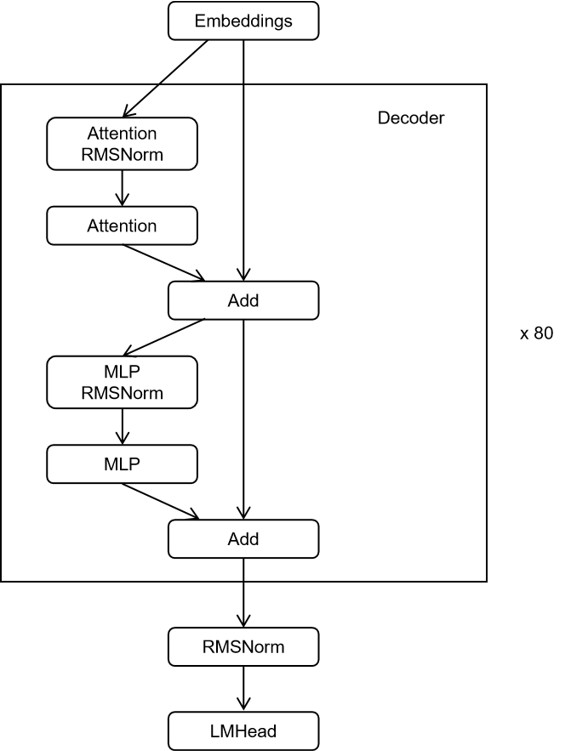}
\caption{Decoder-only architecture overview}
\label{fig:decoder}
\end{figure}


\subsection{Operator Fusion}
Operator fusion merges lightweight memory-bound operators (e.g., normalization, activation) with compute-bound operators (e.g., matrix multiplication, convolution) to execute them together. This approach improves overall compute efficiency and reduces memory footprint and scheduling overhead in deep models. The core objective is to maximize on-chip data locality and minimize accesses to off-chip DRAM, thereby enhancing edge inference performance \cite{11177548}.

From an implementation perspective, operator fusion often models a neural network as a DAG and partitions it into fuseable subgraphs. In these subgraphs, outputs of upstream operators are directly consumed by downstream operators from on-chip caches, avoiding unnecessary intermediate reads/writes \cite{alwani2016fused}. The literature explores a variety of strategies: Optimus \cite{cai2022optimus} uses bottom-up dynamic programming and LUT-based pruning to reduce the search space for off-chip traffic; TileFlow \cite{zheng2023tileflow} combines genetic algorithms with MCTS to quantify data movement and optimize resource allocation when searching for optimal fusion plans.

\label{TT-metalium}
\subsection{Tensix Core}
The Tensix architecture \cite{tenstorr70:online} was proposed by Tenstorrent and is a heterogeneous, scalable AI processor design. Tenstorrent is a semiconductor company focused on RISC\textendash V CPUs and AI accelerators, offering chips centered around Tensix and accelerators such as Wormhole \cite{Wormhole18:online}, together with the TT\textendash metal\cite{tenstorr91:online} low-level programming paradigm and operator library. Unlike traditional GPUs, Tensix decouples compute, control, and communication, and uses a 2D mesh interconnect to form a scalable, dataflow-oriented computation network on chip. The network comprises specialized nodes: Tensix cores serve as compute nodes, complemented by memory nodes, chip management/control nodes, and high-bandwidth communication nodes to improve parallelism and dataflow efficiency.

\begin{figure}[htbp]
\centering
\begin{subfigure}[c]{0.53\textwidth}
\includegraphics[width=\textwidth]{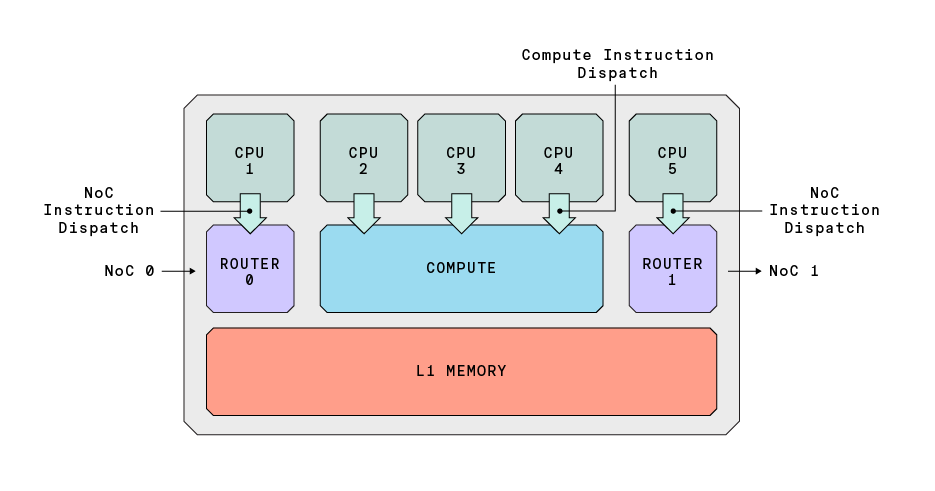}
\caption{Tensix core architecture\cite{ttmetalM8:online}}
\label{fig:tensix_core}
\end{subfigure}
\hfill
\begin{subfigure}[c]{0.46\textwidth}
\includegraphics[width=\textwidth]{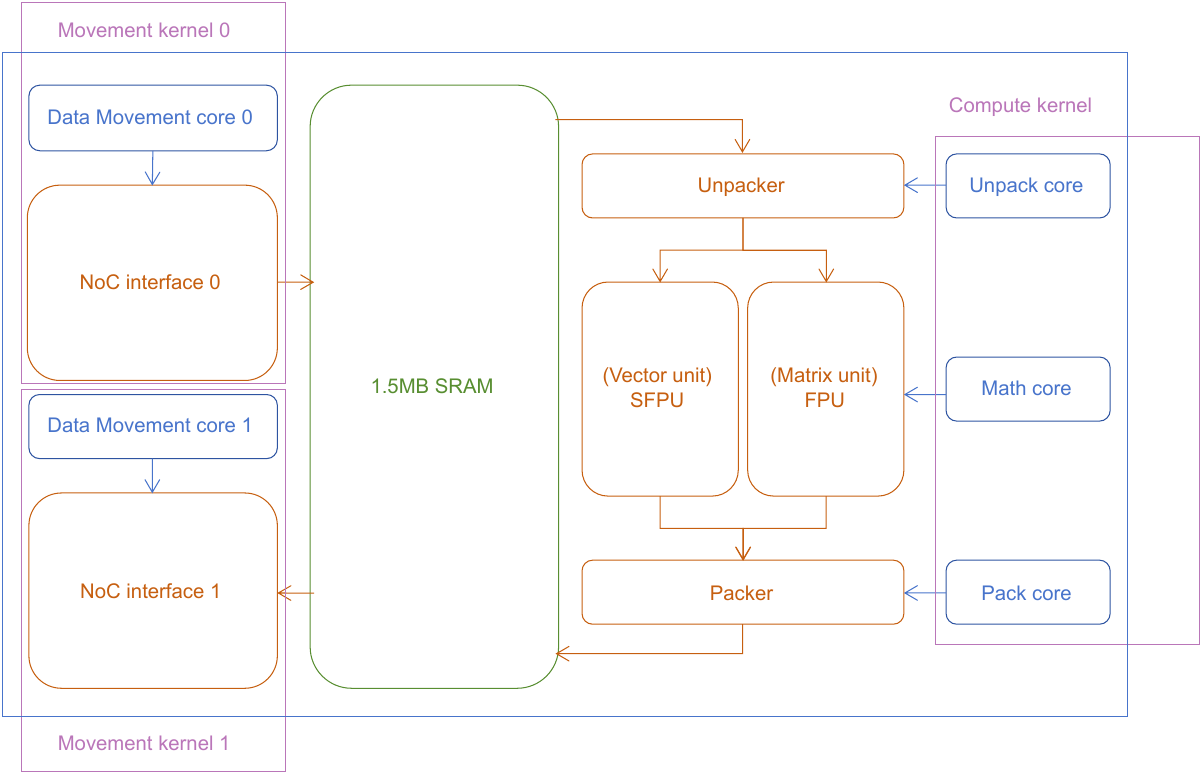}
\caption{Tensix Kernel}
\label{fig:tensix_kernel}
\end{subfigure}
\caption{Tensix architecture overview}
\label{fig:tensix_architecture}
\end{figure}

The Tensix core is the compute unit of the Tensix processor (Fig.~\ref{fig:tensix_core}). Each core contains five small RISC\textendash V control cores \cite{waterman2014risc} (“baby cores”), specialized hardware units for matrix operations (FPU), specialized units for vector operations (SFPU), and 1.5\,MB local SRAM. Typical dataflow: data is delivered to the core via the on-chip NoC, unpacked, processed by the specialized compute units, repacked, and then sent via the NoC to DRAM or other Tensix cores.

As shown in Fig.~\ref{fig:tensix_kernel}, the baby cores handle instruction scheduling and control flow in parallel, issuing commands to the FPU/SFPU to execute computation (instead of running compute directly on the RISC\textendash V cores). The chip provides two independent NoC interfaces circling the die in opposite directions, allowing parallel read/write operations that overlap with computation and form an efficient producer–consumer pipeline.

\subsection{TT-metalium}
TT-metalium \cite{tenstorr91:online} is Tenstorrent’s low-level kernel programming paradigm for implementing custom AI computations on Tensix processors. Its API is OpenCL-like, providing C++ interfaces for high-level operations and direct hardware control, enabling access to the FPU, SFPU, NoC interfaces, and SRAM to write highly optimized kernels with fine-grained control.
TT-metalium adopts the Single Program, Multiple Data (SPMD) execution model \cite{Singlepr87:online}: multiple processing units run the same program over different data shards. On Tensix, a core mesh (common configurations are 64/128 cores) cooperatively processes different blocks of the same computation using SPMD.

As shown in Fig.~\ref{fig:tensix_kernel}, TT-metalium divides the core compute pipeline into three kernel types and coordinates them using circular buffers in the core’s SRAM to implement producer–consumer parallelism. The three kernels fully overlap in time to form an efficient pipeline:
\begin{itemize}
\item Reader kernel: controlled by RISC\textendash V CPU0, using double buffering to asynchronously fetch data from DRAM into SRAM;
\item Compute kernel: controlled by RISC\textendash V CPU1–3, driving data from SRAM to FPU/SFPU for computation and writing results back to SRAM;
\item Writer kernel: controlled by RISC\textendash V CPU4, asynchronously writing results from SRAM back to DRAM.
\end{itemize}

The Tensix core architecture and TT-metalium paradigm have been widely applied to a range of workloads, including solving Laplace’s equation \cite{brown2024accelerating}, fast Fourier transforms \cite{10.1007/978-3-032-07612-0_46}, matrix multiplication \cite{pizzini2025assessing}, attention mechanisms \cite{thuning2024attention}, and high-performance computing \cite{almerol2025accelerating}.

\section{Method}

\subsection{Operator Description}

\subsubsection{RMSNorm Operator}
RMSNorm is a commonly used normalization layer in Transformer models that normalizes input features to prevent gradient vanishing or explosion. Its computation is
\[
\mathrm{RMSNorm}(x;\gamma) \;=\; \frac{x}{\sqrt{\mathrm{E}[x^2] + \epsilon}} \, \odot \, \gamma,
\]
where $\epsilon$ is a small constant, $\gamma$ is a learnable parameter, and $\mathrm{E}[x^2]$ is the mean-square of the input features.
\par In modern decoder-only LLMs (e.g., Llama, Qwen), RMSNorm appears at the start of each residual block as a pre-normalization layer, before self-attention and before the FFN. It stabilizes the numerical range of subsequent linear projections and nonlinear mappings, improving convergence and robustness in training and inference.

\subsubsection{MatMul Operator}
MatMul is the fundamental linear transform primitive in Transformers; its standard form is GEMM:
\[
C \;\leftarrow\; \alpha\,A B \;+\; \beta\,C,\quad A\in\mathbb{R}^{M\times K},\; B\in\mathbb{R}^{K\times N},\; C\in\mathbb{R}^{M\times N}.
\]
\par In the decoder module, the input $X\in\mathbb{R}^{T\times d_{\mathrm{model}}}$ is projected to
\[
Q = X W_Q,\quad K = X W_K,\quad V = X W_V,\quad W_{\{\!Q,K,V\!\}}\in\mathbb{R}^{d_{\mathrm{model}}\times (n_{\mathrm{h}} d_{\mathrm{h}})},
\]
and rearranged as $Q,K,V\in\mathbb{R}^{n_{\mathrm{h}}\times T\times d_{\mathrm{h}}}$. The FFN contains
\[
U = X W_{\mathrm{up}},\quad G = X W_{\mathrm{gate}},\quad Y = \phi(G)\odot U,\quad Z = Y W_{\mathrm{down}},
\]
where $W_{\mathrm{up}}\in\mathbb{R}^{d_{\mathrm{model}}\times d_{\mathrm{ff}}}$ and $W_{\mathrm{down}}\in\mathbb{R}^{d_{\mathrm{ff}}\times d_{\mathrm{model}}}$. These operators account for the majority of FLOPs and are compute-bound, alternating over time with memory-bound operators such as normalization and activation. To improve edge throughput, we fuse the preceding RMSNorm with the above projection matmuls back-to-back. The normalized output is consumed in place in on-chip SRAM, reducing scheduling and off-chip access while forming an efficient producer-consumer pipeline with NoC dataflow.

\subsection{Single-Tensix Operator Fusion}
\begin{figure}[htbp]
\centering
\begin{subfigure}[c]{0.99\textwidth}
\centering
\includegraphics[width=\textwidth]{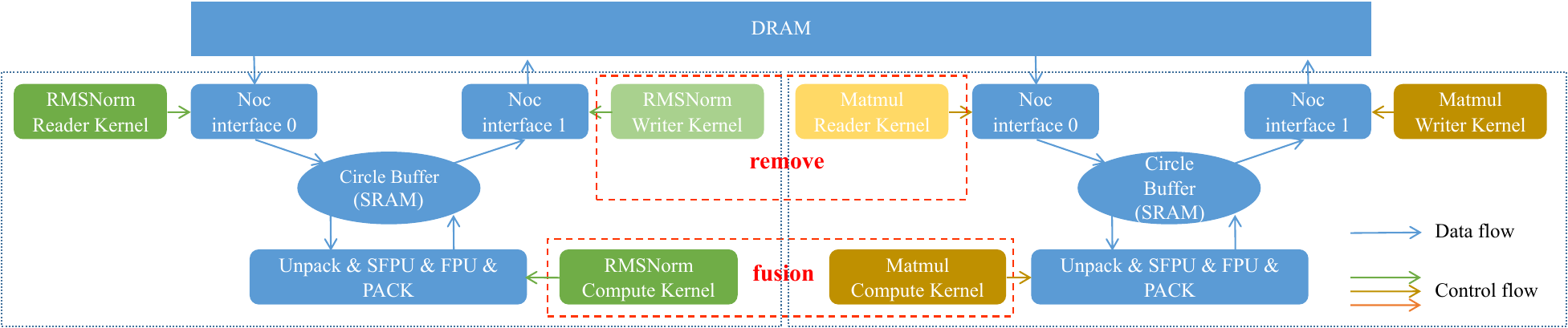}
\caption{Fusion Pre}
\label{fig:single_core_pre}
\end{subfigure}

\vspace{1em}

\begin{subfigure}[c]{0.65\textwidth}
\centering
\includegraphics[width=\textwidth]{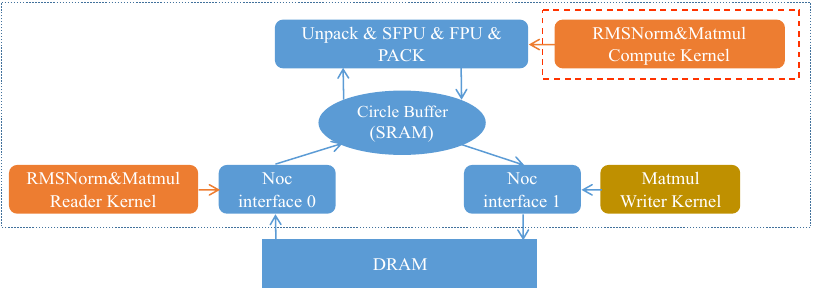}
\caption{Fusion After}
\label{fig:single_core_after}
\end{subfigure}
\caption{Single-Tensix operator fusion illustration}
\label{fig:single_core}
\end{figure}

Fig.~\ref{fig:single_core_pre} shows the standard programming paradigm on Tensix: each standalone operator reads data from shared DRAM into on-chip SRAM, performs computation on SFPU/FPU, and writes results back to DRAM (see Sec.~\ref{TT-metalium}). This “read–compute–write” pattern causes frequent off-chip movement of intermediate data, incurring significant memory latency and scheduling overhead that limits overall efficiency.

To address this, we propose a single-core operator fusion strategy that eliminates DRAM reads/writes of intermediate results via on-chip pass-through: cancel the writeback of the preceding operator (RMSNorm) and the subsequent read of the next operator (matmul). The fused computation is
\[
C \;=\; \mathrm{RMSNorm}(x;\gamma)\,\cdot\, B,
\]
where $x$ is the activation block, $\gamma$ the normalization scale, and $B$ the weight matrix. To ensure efficient hardware realization, we follow these principles:
\begin{itemize}
\item Avoid data movement: keep RMSNorm outputs in on-chip SRAM and feed directly into matmul to avoid unnecessary DRAM transfers and format conversions, reducing inter-operator scheduling and sync overhead.
\item Core-grid alignment: ensure RMSNorm and matmul use identical core-grid partitions so data naturally resides on the correct cores without cross-core migration.
\item Weight reuse: in batched settings, read weights from DRAM once and reuse across multiple activation requests to improve off-chip bandwidth efficiency.
\end{itemize}
The fused workflow (Fig.~\ref{fig:single_core_after}) reduces off-chip round trips via on-chip pass-through and improves bandwidth efficiency and pipeline parallelism through grid alignment and weight reuse.

\subsection{Multi-Tensix Operator Fusion}

\begin{figure}[htbp]
\centering
\includegraphics[width=0.8\textwidth]{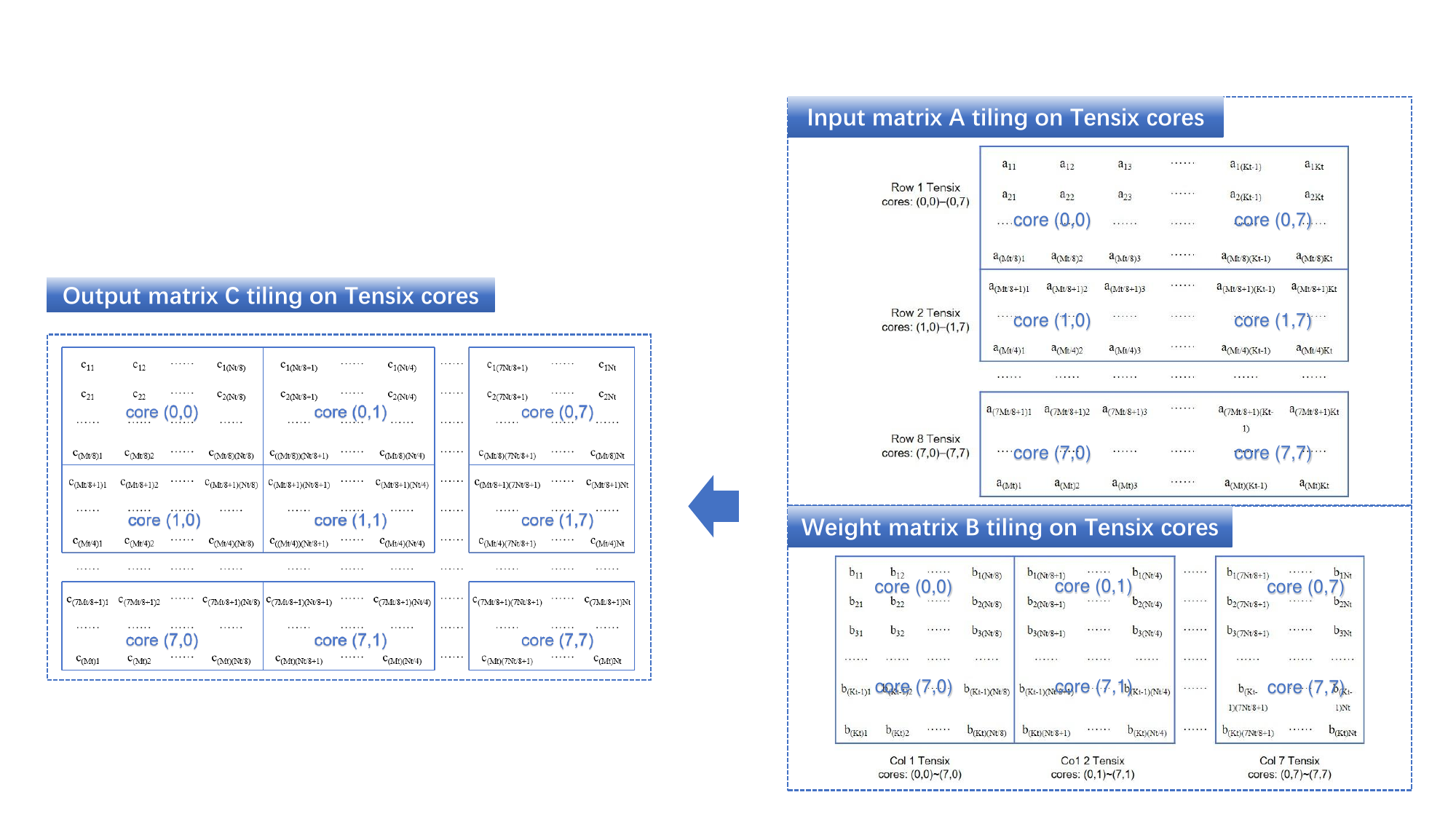}
\caption{Multi-Tensix operator fusion illustration}
\label{fig:multi_core}
\end{figure}

Due to the compute limits of a single Tensix core, large-scale deep learning tasks typically require multiple cores cooperating in parallel. Fig.~\ref{fig:multi_core} shows the fusion strategy in a multi-core setting: partition the computation and map it onto a 2D core mesh. Each core independently performs RMSNorm and matmul for its assigned input tiles, storing intermediate results in local SRAM to minimize inter-core migration.

The key to multi-core parallelism is correctly handling data dependencies. For matmul, Fig.~\ref{fig:multi_core} adopts an output-tiling strategy that maps $C$ uniformly to the core mesh. Algebraically, cores in the same row share row tiles of $A$, while cores in the same column share column tiles of $B$. This property underpins efficient data distribution and parallel computation, enabling cores to execute local work independently and in parallel.

For RMSNorm, since cores in the same row later share row tiles of $A$, all cores in a row must apply identical normalization. To reduce intra-row communication while keeping SPMD consistent, we use compute redundancy: each core in the row performs the same normalization.

\subsection{Multi-Tensix Multicast Acceleration}

While the above partitioning clarifies task layout, it also introduces a new memory bottleneck: multiple cores in the same row or column may simultaneously request identical input or weight tiles from DRAM, causing bandwidth contention and congestion. Under edge constraints with limited off-chip bandwidth, this leads to queueing and arbitration conflicts; pipeline overlap can no longer hide memory latency and throughput drops.

\begin{figure}[htbp]
\centering
\includegraphics[width=0.9\textwidth]{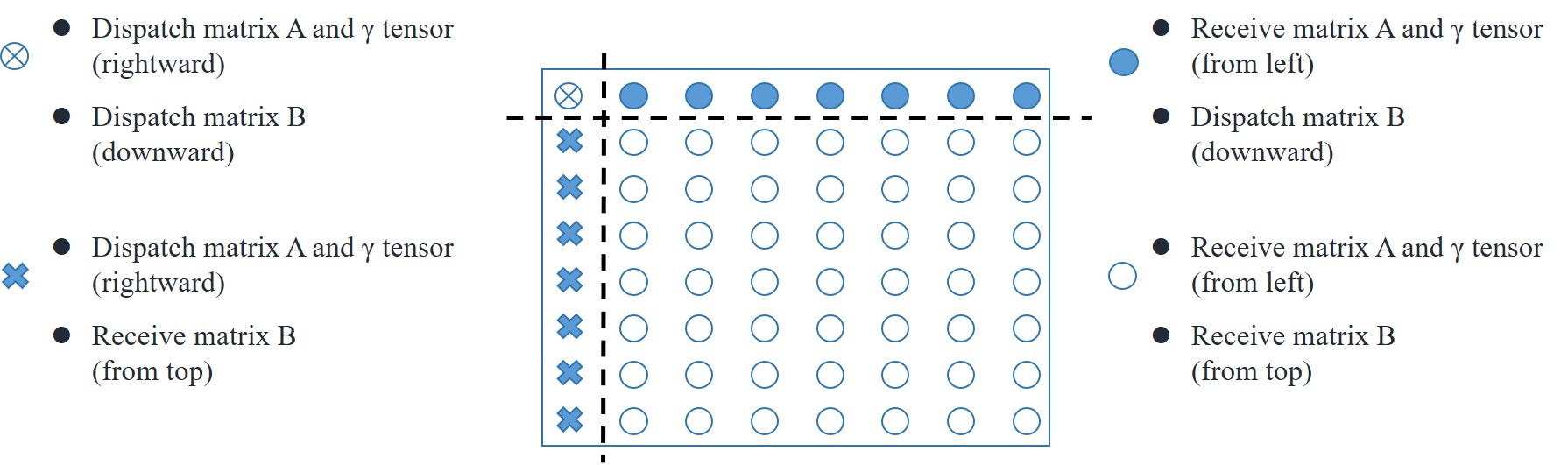}
\caption{Multicast acceleration illustration}
\label{fig:mcast}
\end{figure}

We leverage a NoC-based multicast mechanism grounded in the data-sharing properties of matmul. As shown in Fig.~\ref{fig:mcast}, we adopt a “row master” (leftmost core) in each row to read $A$ and $\gamma$ and multicast horizontally; and a “column master” (topmost core) in each column to read $B$ and multicast vertically.

Row masters fetch $A$ and $\gamma$ row tiles from DRAM into local SRAM circular buffers; upon completing a tile, they trigger horizontal multicast (left$\rightarrow$right). Row followers receive and write into local SRAM. Column masters fetch $B$ column tiles into another buffer; upon completing a tile, they trigger vertical multicast (up$\rightarrow$down). Column followers receive and write into local SRAM.

Each core consumes data locally using a producer–consumer pipeline: once the corresponding $A/\gamma$ and $B$ tiles arrive, it immediately performs the fused RMSNorm+matmul without writing intermediate results to DRAM. To ensure tile order consistency and avoid congestion, a lightweight token or fence synchronization controls buffer read/write cadence while maintaining overlap of compute and communication.
This mechanism transforms dense DRAM–core communication into efficient core–core on-chip communication, substantially reducing DRAM bandwidth pressure, improving memory access efficiency, and exploiting spatial data locality.

\section{Experiment Results}
We use the Tenstorrent Wormhole N300 accelerator card as our evaluation platform. The device connects via PCIe to a ThinkPad X1 laptop running openKylin SP2, forming a representative edge inference environment. The N300 integrates two Tensix chips for a total of 128 Tensix cores, 24\,GB GDDR6, and 192\,MB SRAM. Its peak performance reaches 466 TFLOPS (FP8) and 131 TFLOPS (FP16). Notably, since Wormhole employs a dual-chip design, we can configure it to a single-chip mode (i.e., N150 with 64 Tensix cores) to validate algorithm adaptability across different hardware scales.

To comprehensively evaluate our fusion strategy, we benchmark against Qwen2.5-0.5B, Qwen3-0.6B, and Qwen3-4B. Qwen2.5 is an open-source LLM built on the decoder-only Transformer architecture, integrating SwiGLU activation, RoPE positional encoding, and RMSNorm; it performs well in code generation and mathematical reasoning. Qwen3 iterates on the series with parameter and structural optimizations that reduce inference cost while maintaining high performance. All these models exhibit typical compute–memory interleaving characteristics, making them ideal for validating fusion performance.

For performance analysis, we use Tenstorrent's official ttnn-visualizer \cite{tenstorr97:online} for fine-grained timing statistics. It records each operator's device execution time and the scheduling gap between operators. Our fusion strategy aims to minimize or eliminate these inter-operator gaps, thereby significantly increasing overall inference throughput.

For accuracy, we measure numerical consistency between fused and baseline implementations using the Pearson Correlation Coefficient (PCC). Let the baseline output be $X=\{x_i\}_{i=1}^{N}$ and the fused output be $Y=\{y_i\}_{i=1}^{N}$; PCC is defined as
\[
r=\frac{\sum_{i=1}^{N}(x_i-\bar{x})(y_i-\bar{y})}{\sqrt{\sum_{i=1}^{N}(x_i-\bar{x})^2}\;\sqrt{\sum_{i=1}^{N}(y_i-\bar{y})^2}}
\]
with range $[-1,1]$: values closer to $1$ indicate stronger linear correlation. Insensitive to translation and scaling, PCC is well-suited for comparing numerical consistency across implementations.

\begin{table}[htbp]
\centering
\caption{Performance comparison of fused kernels across different model versions.}
\label{tab:performance_comparison}
\begin{tabular}{ccccccc}
\toprule
\textbf{Tensix Cores} & \textbf{Model} & \textbf{Size} & \textbf{Attn} & \textbf{MLP} & \textbf{Decoder} & \textbf{Accuracy} \\
\textbf{(Count)} & \textbf{} & \textbf{} & \textbf{Latency $\downarrow$} & \textbf{Latency $\downarrow$} & \textbf{Latency $\downarrow$} & \textbf{(PCC)} \\
\midrule
64 & Qwen2.5-0.5B & 1 GB & 37.44\% & 12.04\% & 7.91\% & 99.94\% \\
64 & Qwen3-0.6B   & 1.2 GB & 18.53\% & 5.66\%  & 4.58\% & 99.57\% \\
128 & Qwen3-4B    & 8 GB & 10.63\% & 15.89\% & 3.58\% & 98.75\% \\
\bottomrule
\end{tabular}
\end{table}

Table~\ref{tab:performance_comparison} details latency and accuracy across model versions and hardware configurations (number of Tensix cores) under fusion. Results show significant latency reductions for key modules. For Qwen2.5-0.5B at 64 cores, attention and MLP latencies decrease by 37.44\% and 12.04\%, respectively, yielding a 7.91\% reduction in single decoder-layer latency. Even for the larger Qwen3-4B at 128 cores, decoder-layer latency improves by 3.58\%. Crucially, PCC remains above 98.75\% in all settings (up to 99.94\%), indicating that our fusion strategy substantially improves efficiency and hardware utilization while strictly preserving inference accuracy.

\section{Conclusion}
\noindent
We propose a locality-driven operator fusion strategy for edge inference under the memory hierarchy constraints of the Tenstorrent Tensix architecture and the compute–memory alternation in decoder-only Transformers. The core idea is to execute RMSNorm with self-attention QKV projections and RMSNorm with the first FFN matmul back-to-back in on-chip SRAM, combined with 2D mesh mapping and NoC row/column multicast to reduce DRAM reads/writes of intermediate results and bandwidth contention, thereby improving throughput and energy efficiency.

On Wormhole N300 with Qwen2.5-0.5B, Qwen3-0.6B, and Qwen3-4B, attention and MLP latencies decrease by up to 37.44\% and 15.89\%, respectively, and single decoder-layer latency decreases by up to 7.91\%. PCC remains above 98.75\% throughout. These results validate that, under numerical stability and accuracy guarantees, our method significantly alleviates off-chip bottlenecks and improves end-to-end inference efficiency.

Overall, our work jointly optimizes at the operator level and the on-chip dataflow level: single-core pass-through fusion tightly couples memory-bound and compute-bound operators, while multi-core 2D mapping with row/column multicast reduces cross-core duplicate reads and bandwidth contention. Together these shorten the Op-to-Op gap, make pipeline overlap more effective, and increase hardware utilization.

The method does not depend on model-specific details and generalizes well to mainstream decoder-only LLMs with RMSNorm, RoPE, and SwiGLU. It yields stable gains across core scales and parameter sizes while maintaining numerical consistency under strict PCC criteria, providing a practical engineering path for deploying medium-to-large LLMs in bandwidth-constrained edge environments.


%
%
%
%





\bibliographystyle{splncs04} 
\bibliography{mybibliography} 
\end{document}